\def\isarxiv{1} 

\ifdefined\isarxiv
\documentclass[11pt]{article}

\usepackage[numbers]{natbib}

\else
\documentclass{uai2025}
\usepackage[american]{babel}
\usepackage{natbib}

\fi

\usepackage{amsmath}
\usepackage{amsthm}
\usepackage{amssymb}
\usepackage{algorithm}
\usepackage{subfig}
\usepackage{algpseudocode}
\usepackage{graphicx}
\usepackage{grffile}
\usepackage{wrapfig,epsfig}
\usepackage{url}
\usepackage{xcolor}
\usepackage{epstopdf}

\usepackage{bbm}
\usepackage{dsfont}

\allowdisplaybreaks

\ifdefined\isarxiv

\usepackage{tikz}
\usepackage{hyperref}  
\hypersetup{colorlinks=true,citecolor=blue,linkcolor=blue} 
\usetikzlibrary{arrows}
\usepackage[margin=1in]{geometry}

\else

\usepackage{microtype}
\usepackage{hyperref}
\definecolor{mydarkblue}{rgb}{0,0.08,0.45}
\hypersetup{colorlinks=true, citecolor=mydarkblue,linkcolor=mydarkblue}

\fi
 
\graphicspath{{./figs/}}

\theoremstyle{plain}
\newtheorem{theorem}{Theorem}[section]
\newtheorem{lemma}[theorem]{Lemma}
\newtheorem{definition}[theorem]{Definition}

\newtheorem{assumption}[theorem]{Assumption}

\newtheorem{example}[theorem]{Example}

\newcommand{\wh}{\widehat}
\newcommand{\wt}{\widetilde}

\newcommand{\N}{\mathcal{N}}
\newcommand{\R}{\mathbb{R}}

\renewcommand{\d}{\mathrm{d}}

\renewcommand{\hat}{\wh}

\DeclareMathOperator*{\E}{{\mathbb{E}}}

\makeatletter
\newcommand*{\RN}[1]{\expandafter\@slowromancap\romannumeral #1@}
\makeatother

\usepackage{lineno}

\begin{document}

\ifdefined\isarxiv

\date{}

\title{Theoretical Foundation of Flow-Based Time Series Generation:\\Provable Approximation, Generalization, and Efficiency}
\author{
Jiangxuan Long\thanks{\texttt{ lungchianghsuan@gmail.com}. South China University of Technology.}
\and 
Zhao Song\thanks{\texttt{ magic.linuxkde@gmail.com}. The Simons Institute for the Theory of Computing at the UC, Berkeley.}
\and
Chiwun Yang\thanks{\texttt{ christiannyang37@gmail.com}. Sun Yat-sen University.}
}

\else

\title{Theoretical Foundation of Flow-Based Time Series Generation:\\Provable Approximation, Generalization, and Efficiency}
\author{
Intern Name
}
\maketitle
\fi

\ifdefined\isarxiv
\begin{titlepage}
  \maketitle
  \begin{abstract}

Recent studies suggest utilizing generative models instead of traditional auto-regressive algorithms for time series forecasting (TSF) tasks. These non-auto-regressive approaches involving different generative methods, including GAN, Diffusion, and Flow Matching for time series, have empirically demonstrated high-quality generation capability and accuracy. However, we still lack an appropriate understanding of how it processes approximation and generalization. This paper presents the first theoretical framework from the perspective of flow-based generative models to relieve the knowledge of limitations. In particular, we provide our insights with strict guarantees from three perspectives: {\bf Approximation}, {\bf Generalization} and {\bf Efficiency}. In detail, our analysis achieves the contributions as follows:
\begin{itemize}
    \item By assuming a general data model, the fitting of the flow-based generative models is confirmed to converge to arbitrary error under the universal approximation of Diffusion Transformer (DiT).
    \item Introducing a polynomial-based regularization for flow matching, the generalization error thus be bounded since the generalization of polynomial approximation.
    \item The sampling for generation is considered as an optimization process, we demonstrate its fast convergence with updating standard first-order gradient descent of some objective.
\end{itemize}

  \end{abstract}
  \thispagestyle{empty}
\end{titlepage}

{\hypersetup{linkcolor=black}
\tableofcontents
}
\newpage

\else
\begin{abstract}

Recent studies suggest utilizing generative models instead of traditional auto-regressive algorithms for time series forecasting (TSF) tasks. These non-auto-regressive approaches involving different generative methods, including GAN, Diffusion, and Flow Matching for time series, have empirically demonstrated high-quality generation capability and accuracy. However, we still lack an appropriate understanding of how it processes approximation and generalization. This paper presents the first theoretical framework from the perspective of flow-based generative models to relieve the knowledge of limitations. In particular, we provide our insights with strict guarantees from three perspectives: {\bf Approximation}, {\bf Generalization} and {\bf Efficiency}. In detail, our analysis achieves the contributions as follows:
\begin{itemize}
    \item By assuming a general data model, the fitting of the flow-based generative models is confirmed to converge to arbitrary error under the universal approximation of Diffusion Transformer (DiT).
    \item Introducing a polynomial-based regularization for flow matching, the generalization error thus be bounded since the generalization of polynomial approximation.
    \item The sampling for generation is considered as an optimization process, we demonstrate its fast convergence with updating standard first-order gradient descent of some objective.
\end{itemize}

\end{abstract}

\fi

\section{Introduction}
Generative models have revolutionized machine learning by enabling the creation of highly realistic and diverse content across various domains. In particular, diffusion-based approaches~\cite{ddpm}, Generative Adversarial Networks~\cite{gan4}, and flow matching methods~\cite{lcb+23} have emerged as powerful tools for data synthesis and augmentation. These methods leverage sophisticated architectures to learn complex probability distributions and transform random noise into structured, meaningful outputs. For example, text-to-image models translate textual descriptions into compelling visual artworks or photographs~\cite{zra23}, while recent advances in text-to-video frameworks produce coherent and temporally consistent video content~\cite{hsg+22}. Discrete flow matching~\cite{grs+24} extends continuous-time flow-based modeling to discrete settings by carefully aligning discrete probability distributions via flexible transformations, thereby broadening the applicability of flow-based generative models to high-dimensional discrete domains such as language and code. As these techniques continue to evolve, the ability of generative models to capture intricate data structures and produce high-quality samples underscores their broadening influence in artificial intelligence research.

Among all these data types, time series data, found in fields like finance, healthcare, and climate science, constitutes a critical yet challenging domain for forecasting and analysis \cite{bj76}. Given its temporal dependency and noisy nature~\cite{bjrl15}, time series poses unique obstacles that often exceed the complexities encountered in static data settings. By establishing the NP-hardness of computing a mean in dynamic time-warping spaces, \cite{bfn20} highlights key computational challenges in time series analysis.
Nonetheless, the powerful capabilities of generative models have proven effective in tackling these challenges, offering promising solutions on time series data. By learning the underlying distribution of time series trajectories, generative approaches can capture both signal and noise components, thereby producing more robust forecasts and generalizations. Indeed, the recent success of GAN~\cite{jks+22}, diffusion~\cite{rssv21,tsse21}, and flow-based models~\cite{zpk+24} in time series highlights their growing appeal, as these tools exhibit strong empirical performance across diverse application scenarios \cite{app1,app2,app3}. Consequently, the burgeoning research on generative models for temporal data generation and forecasting stands at the forefront of machine learning, offering transformative potential for both academia and industry.

Although such generative models show remarkable performance when applied to time series, our theoretical understanding of their success remains limited. Researchers have begun questioning what fundamental principles govern their approximation capabilities and how well they generalize under real-world data conditions \cite{zyll24,fsi+25}. Without a solid theoretical framework, it is difficult to fully trust and optimize these methods, and their reliability in safety-critical domains becomes a concern. While empirical evidence consistently demonstrates their potential, the absence of a rigorous conceptual foundation obscures deeper insights into model selection, hyperparameter tuning, and design strategies. Indeed, bridging this gap between practical efficacy and theoretical clarity is an urgent priority, which motivates our efforts to explore flow-based generative models for time series and provide meaningful error bounds and generalization guarantees.

In this work, we propose a strict framework to analyze the generative models for time series generation, especially the flow-based generative models~\cite{hww+24,yq24}. It involves three parts: 
\begin{itemize}
    \item {\it Approximation.} Theorem~\ref{thm:main_1} confirm that flow-based generative models converge to arbitrary approximation error under the universal approximation capability of DiT in Section~\ref{sec:app}.
    \item {\it Generalization.} Theorem~\ref{thm:main_2} derive bounded generalization error guarantees, leveraging the inherent approximation properties of orthogonal polynomial bases to ensure robustness against noise and distribution shifts in Section~\ref{sec:gen}.
    \item {\it Efficiency.} Theorem~\ref{thm:main_3} in Section~\ref{sec:eff} establishes fast convergence guarantees through gradient descent dynamics, demonstrating that our framework achieves efficient generation while maintaining theoretical stability.
\end{itemize}

\paragraph{Roadmap.} In Section~\ref{sec:related}, we review relevant related work. Section~\ref{sec:preli} introduces key background concepts and the problem setup. In Section~\ref{sec:flow}, we present the framework for time series generation using flow matching. Section~\ref{sec:app} discusses the approximation results, while Section~\ref{sec:gen} covers generalization results. Section~\ref{sec:eff} examines efficiency results. Finally, we conclude our paper in Section~\ref{sec:conclusion}.
\section{Related Work}\label{sec:related}
We briefly introduce some topic that are closely related to this work: Generative Models, State Space Models, Understanding Transformer-Based Models, and Time Series Forcasting.

{\bf Generative Models. }
Generative models have emerged as a powerful framework for learning complex data distributions, encompassing methods such as Variational Autoencoders (VAEs) \cite{vae1,vae2}, Generative Adversarial Networks (GANs) \cite{gan1,gan2,gan3,gan4}, and diffusion-based approaches \cite{diffusion1} that iteratively refine noisy samples. VAEs introduce a latent-variable formulation with an encoder-decoder architecture to learn a smooth latent space, while GANs employ a minimax game between generator and discriminator to capture sharp data distributions. Recent diffusion approaches, such as Denoising Diffusion Probabilistic Models (DDPM) \cite{ddpm}, progressively destroy data by adding noise and then reverse the process via learned denoising steps. Score-based methods \cite{score1,score2} generalize this process by estimating the gradient (score) of the data density to generate samples through stochastic differential equations. Normalizing flows \cite{nf1,nf2} take an alternative route by constructing invertible transformations with tractable Jacobians, enabling exact likelihood computation. More recently, novel paradigms such as flow matching \cite{lcb+23} and rectified flow \cite{lgl23} have emerged, aiming to simplify sampling via direct trajectory-based transformations. In parallel, advancements in Diffusion Probabilistic Model (DPM) solvers \cite{dpm1,dpm2} further optimize the sampling process, reducing computational overhead while preserving generative fidelity. Collectively, these developments highlight a vibrant research landscape, where systematic improvements and new theoretical insights continue to push the boundaries of generative modeling~\cite{cgl+25_homo,cll+25_var,gll+25,ccl+25,gkl+25,cgh+25,lss+25_high}.

{\bf Understanding Transformers-Based Models.} 
Understanding Transformer-based models involves tackling the quadratic complexity of their attention mechanism, prompting innovations such as sparse attention, low-rank approximations, and kernel-based methods to reduce computational demands and boost scalability \cite{vsp+17,hci+21,kkf+23,fa23,llss24_sparse,rsw16,llr16,eyp+22,zk24,hsk+24,ckns20,lz20,dls+25_pca,zhdk23,lsss24}. Some approaches focus on deriving low-rank representations of the attention matrix, enabling near-linear time acceleration for single-layer and multi-layer Transformers \cite{aa22,as23,as24,as24_iclr,lss+24}. Others, including Linearizing Transformers, Hopfield Models, and PolySketchFormer, rely on architecture modifications and implementation optimizations to further enhance performance \cite{hcl+24,kmz23,cgl+25_scaling,hwl24,cll+25_kv,hyw+23,zbkr24,lll+25_loop,hcw+24,xhh+24,mvk+24,whl+24,whhl24,hlsl24}. System-level techniques, such as FlashAttention and block-wise parallel decoding, also bolster efficiency for longer input sequences, broadening real-world applicability \cite{dfe+22,d23,sbz+24,ssu18,cqt+23,pqfs23,lls+24_io,lssy24,smn+24}. Beyond these efficiency gains, numerous strategies have emerged to adapt large language models (LLMs) for specialized tasks—examples include adapters, calibration methods, multitask fine-tuning, prompt tuning, scratchpad techniques, instruction tuning, symbol tuning, black-box tuning, reinforcement learning from human feedback, and chain-of-thought reasoning \cite{zmc+24,vnr+23,eyp+22,zhz+23,owj+22,kls+25,ghz+23,zzj+23b,zzj+23a,chl+22,mkd+22,lac+21,zjk+23,naa+21,jla+23,sdj+23,ssy+22,cpp+23,zwf+21,ll21,ksk+22,gfc+21a,wws+22}. Recent research on tensor Transformers, acceleration techniques, and related advancements further refines our understanding of these models, guiding continued optimization and novel applications \cite{dsy24a,alsy24,hlsl24,hsk+24,csy24b,whhl24,hyw+23,ssz+24_pruning,cll+25_fair,zha24,zxf+24,szz24,hwsl24,kll+25,lssz24_dp,lls+24_grok,gsx23,dsy24b,dswy22_coreset,qsw23_submodular,lsy24,hcw+24,csy24a,lls+24_tensor,llss24_sparse,lls+24_prune,lls+25_graph,cll+24_rope,cll+24_ssm,lls+24_dp_je,dlg+22,whl+24,lls+24_conv,lssy24,dsy24sparse,ssz23_tradeoff,hwl24,sy23_des,klsz24_sample_tw,as24_rope,smn+24,lls+24_io,chl+24_rope_grad,cls+24,ssx23_ann,gms23_exp_reg,xhh+24,hcl+24,lssz24_tat,zly+25,sht24,ssz+24_dit,xsl24,llsz24_nn_tw}.

{\bf Time Series Forecasting and Imputation. }
Time series forecasting has evolved significantly from classical statistical models, such as ARIMA \cite{bj76} and ETS \cite{g85}, to sophisticated deep learning techniques that capture complex nonlinear and long-term dependencies. Early neural network approaches, notably recurrent architectures like the Long Short-Term Memory (LSTM) \cite{lstm} and Gated Recurrent Units (GRU) \cite{gru}, demonstrated considerable success in modeling sequential data, paving the way for sequence-to-sequence frameworks \cite{seq2seq}.  
Recent advances in time series forecasting have spurred a wide range of innovative methodologies that tackle long-range dependencies, interpretability, and generative performance. For instance, Transformer-based approaches have been enhanced by novel attention mechanisms that preserve temporal correlations, as demonstrated in~\cite{klk+24}, and by deformable architectures that mitigate the limitations of patching, as shown in~\cite{lw24}. Complementing these efforts, segmentation strategies that convert time series into subseries-level patches have been proposed for efficient long-term forecasting in~\cite{nns+22}. In parallel, decomposition methods have been leveraged to balance model complexity and capability, achieving robust performance with dramatically fewer parameters as detailed in~\cite{dyy+24}, while structured matrix bases have been employed to yield interpretable multivariate dynamics in forecasting models~\cite{clcl24}. On another front, diffusion models have emerged as a potent tool for both forecasting and generative modeling. For example, retrieval-augmented diffusion frameworks use historical samples to guide denoising processes during prediction~\cite{lylh24}, and image-transform techniques convert time series data into images to harness the strengths of diffusion-based generation~\cite{nbp+24}. Enhancements to the diffusion paradigm include conditional models that inflate high-frequency components to capture extreme events~\cite{gtl24} and adaptive noise scheduling that tailors the diffusion process to the non-stationarity of the data~\cite{llp24}. Moreover, the integration of large language models (LLMs) into time series forecasting has led to promising yet debated approaches. On one hand, autoregressive forecasting frameworks leveraging LLMs have achieved state-of-the-art performance with remarkable efficiency \cite{lqh+24}, while on the other hand, studies have questioned whether the complexity of LLMs is necessary, suggesting that simpler attention layers might suffice \cite{tmg+24}. Further, the incorporation of external event analysis through LLMs has enriched forecasting by aligning news with time series fluctuations \cite{wfq+24}, and tri-level learning frameworks have exploited LLM representations to enhance out-of-distribution generalization \cite{jyj24}. 

\section{Preliminary}\label{sec:preli}

This section introduces the theoretical background we aim to solve in this paper. In detail, we introduce the key notations and definitions for window sizes, pseudoinverses, and other fundamental concepts in Section~\ref{sec:preli:notation}. 
In Section~\ref{sec:preli:problem}, we formally define the time series forecasting and imputation problem by presenting the data model, assumptions on smooth signals and Gaussian noise, and the objective function. 
Finally, in Section~\ref{sec:preli:poly}, we explore polynomial approximation bases, highlighting their orthogonality, positive definiteness, and strong approximation capabilities in modeling time series data.

\subsection{Notations}\label{sec:preli:notation}
We use $[n]$ to denote the set $\{1, 2, \cdots, n\}$. We use $\E[]$ to denote the expectation. We use $\|A\|_F$ to denote the Frobenius norm of a matrix $A \in \R^{n \times d}$, i.e. $\|A\|_F := \sqrt{\sum_{i \in [n]} \sum_{j \in [d]} |A_{i,j}|^2}$. We use $\|\cdot\|_2$ to denote the $\ell_2$ norm of a vector $x \in \R^{d}$, i.e. $\|x\|_2 := \sqrt{ \sum_{i \in [d]} |x_i|^2}$. We use $\|\cdot\|_\infty$ to denote the $\ell_\infty$ norm of a vector $x \in \R^{d}$, i.e. $\|x\|_\infty := \max_{i\in [d]} x_i$. We use positive integer $N_x$ to denote the window size of input data, and positive integer $N_y$ to denote the window size of output data, especially, we have $N_x \gg N_y$ and denote $N := N_x + N_y$. The function $\lambda_{\min}: \R^{d_1 \times d_2} \rightarrow \R$ takes any matrix $A \in \R^{d_1 \times d_2}$ as input and outputs the smallest singular value of matrix $A$. We use $|\cdot|$ to represent the size of a set. We use $e_\tau \in \R^{N}$ to denote the $N$-dimensional one-hot vector with the $\tau$-th entry is $1$ for any $\tau \in [N]$. For any matrix $A \in \R^{d_1 \times d_2}$, we use $A^\dagger \in \R^{d_2 \times d_1}$ to stands for its pseudoinverse. We say a matrix $A$ is positive definite (PD) once its smallest singular value is positive, $\lambda_{\min}(A) > 0$.

\subsection{Problem Definition: Data Model}\label{sec:preli:problem}

We first define the data model of time series: We consider a distribution ${\cal D}$ containing discrete time series in the $N$-dimensional vector form, none of any two time series are equal. 
\begin{definition}[Data model]\label{def:data_model}
    We define the data model of time series as
    \begin{align*}
        {\cal D} = \{f^r\}_{r=1}^{|{\cal D}|} \subset \R^N.
    \end{align*}
    which satisfied that none of any two time series in $\cal D$ are equal.
\end{definition}

Furthermore, we denote input indices set ${\cal I}_x$ and output indices set ${\cal I}_y$ as 
\begin{definition}\label{def:indices_set}
    we define input indices set
    \begin{align*}
        |{\cal I}_x| := N_x
    \end{align*}
    and output indices set 
    \begin{align*}
        {\cal I}_y := [N]-{\cal I}_x.
    \end{align*}
\end{definition}

Hence, we split each $f \in {\cal D}$ into input series $f_x \in \R^{N_x}$ and output series $f_y \in \R^{N_y}$. 
\begin{definition}[Time series]\label{def:f}
    Let ${\cal D}$ be defined in Definition~\ref{def:data_model}. Let ${\cal I}_x$ and ${\cal I}_y$ be defined in Definition~\ref{def:indices_set}. Let the observation matrix $M({\cal I}) := [e_\tau^\top]_{\tau \in {\cal I}} \in \R^{|{\cal I}| \times N}$.
    We define input series $f_x \in \R^{N_x}$ and output series $f_y \in \R^{N_y}$ by splitting each time series $f \in {\cal D}$ such that 
    \begin{align*}
        f := [f_x,f_y].
    \end{align*}
    Also, we have
    \begin{align*}
        f_x = M({\cal I}_x) \cdot f ~~\mathrm{and}~~  f_y = M({\cal I}_y) \cdot f.
    \end{align*}
\end{definition}

The main goal of this paper, both time series forecasting and imputation, is to find an algorithm $F \in {\cal F} : \R^{N_x} \rightarrow \R^{N_y}$ (from some function class, taking $f_x$ as input and outputs predictive time series) that the following optimization problem
\begin{definition}\label{def:problem}
    Let ${\cal F} := \{ f ~|~ f: \R^{N_x} \rightarrow \R^{N_y} \}$. Let $F \in {\cal F}$.
    We define our optimization problem as follows:
    \begin{align*}
    \min_{F \in {\cal F}} \E_{f \in {\cal D}}[ \| F(f_x) - f_y \|_2^2 ].
\end{align*}
\end{definition}

Further, we make the following assumptions. First, we assume that each time series sample from the signal with noise.
\begin{assumption}[Time series sample]\label{asp:sample_f}
    Let $g: \R^{\geq 0} \rightarrow \R$ be the signal. Let $\xi_\tau$ be the noise which satisfies Assumption~\ref{asp:gaussian_noise}. Let $\Delta > 0$ be sampling step size. Let $\cal D$ be defined in Definition~\ref{def:data_model}.
    We assume that each time series $f \in {\cal D}$ sampled from 
    \begin{align*}
        f = g(\tau \cdot \Delta) + \xi_\tau, \forall \tau \in [N].
    \end{align*}
\end{assumption}

We assume signal $g$ is Lipschitz smooth with constant $L_0 > 0$.
\begin{assumption}[$L$-smooth and]\label{asp:smooth}
    Let $g: \R^{\geq 0} \rightarrow \R$ be the signal.
    We assume signal $g$ is Lipschitz smooth with constant $L_0 > 0$, that is 
    \begin{align*}
        | g(t) - g(t') | \leq L_0\cdot |t - t'|
    \end{align*}
    for any $t, t' \geq 0$.
\end{assumption}

Moreover, we assume $g(\tau \cdot \Delta)$ is bounded.
\begin{assumption}[bounded signal $g$]\label{asp:bound_g}
    Let $\Delta > 0$ be sampling step size. Let $g: \R^{\geq 0} \rightarrow \R$ be the signal. 
    We assume that
    \begin{align*}
        g(\tau \cdot \Delta) \leq O(\sqrt{N})
    \end{align*}
    for any $\tau \in [N]$.
\end{assumption}

We assume noise $\xi_\tau$ is sampled from a zero-mean and $v$-variance Gaussian distribution.
\begin{assumption}[Noise is Gaussian]\label{asp:gaussian_noise}
    Let $\xi_\tau$ be the noise. Let $v > 0$ be the variance.
    We sample the noise from the $v$-variance Gaussian distribution, i.e.,
    \begin{align*}
        \xi_\tau \sim \N(0,v)
    \end{align*}
\end{assumption}

Here we transform addressing Definition~\ref{def:problem} into addressing to minimize the function regret utilizing neural network $F_\theta \in \R^{N_x} \rightarrow \R^{N_y}$ with parameters $\theta$. We aim to minimize the regret on both in-distribution and out-of-distribution data.
\begin{definition}\label{def:trans_loss}
    Let the optimization problem be defined in Definition~\ref{def:problem}. Let $F_\theta \in \R^{N_x} \rightarrow \R^{N_y}$ be the neural network with parameter $\theta$. Let $F^* := \underset{F \in {\cal F}}{\arg \min} \E_{f \in {\cal D}}[ \| F(f_x) - f_y \|_2^2 ]$ be the optimal fitted function.
    We define the transformed optimization problem which regret on both in-distribution and out-of-distribution data as follows:
    \begin{align*}
        \min_{\theta} \E_{f_x \in \R^{N_x}}[\| F_\theta(f_x) - F^*(f_x) \|_2^2].
    \end{align*}
\end{definition}

\subsection{Polynomial Approximation}\label{sec:preli:poly}

In the following, we discuss a group of specific polynomial bases. Since the strong approximating ability to differentiable functions like the Fourier approximation (usually converging to an arbitrary error with a sufficiently high order), previous works~\cite{yq24,hww+24} apply such approach as some regularization method that provides the model with prior knowledge. In the range of this paper, we define a sequence of specific orthogonal polynomial bases as
\begin{definition}[Orthogonal polynomial bases]\label{def:poly_base}
    Let $n$ be the number of orders of the polynomials. We define the orthogonal polynomial bases $P$ as
    \begin{align*}
        P := \begin{bmatrix}
            P_1, P_2, \cdots, P_n
            \end{bmatrix} \in \R^{N \times n}
    \end{align*}
    where each column $P_i \in R^N$ for any $i \in [n]$ is a polynomial basis.
    It satisfies that
    \begin{itemize}
        \item The degree of $P_i \in \R^N$ for any $i \in [n]$, denotes ${\rm deg}(P_i) = i-1$.
        \item Each polynomial basis is orthogonal due to some measurement $\ell$. Formally, $\langle P_i, P_j\rangle_\ell = 0$.
        \item $P$ is positive definite (PD),  such that $\lambda := \lambda_{\min}(P) > 0$.
        \item The upper bound on $\ell_\infty$ norm of $P$ is $\exp(O(nN))$.
    \end{itemize}
\end{definition}

The approximating capability of polynomial approximation is obvious. To show that, we first introduce a tool from previous work:
\begin{lemma}[Proposition 6 in~\cite{gde+20}]\label{lem:poly_approx_hippo}
    If the following conditions hold:
    Let $f: \R^{\geq 0} \rightarrow \R$ be a differentiable function. Let $g_t := \mathrm{proj}_t (f)$ be its projection at time $t$ with maximum polynomial degree $N-1$. Assume $f$ is $L$-Lipschitz.
    Then we have
    \begin{align*}
        \|f - g_t \|_2 = O(tL/ \sqrt{N}).
    \end{align*}
\end{lemma}

Apply the above lemma, we can show
\begin{lemma}\label{lem:poly_approx}
    Let $g: \R^{\geq 0} \rightarrow \R$ be the signal. Let $f' := [g(\tau \cdot \Delta)]_{\tau=1}^N$ be the sample for some signal $g$, where $\Delta$ is the sample step size.
    Then we have
    \begin{align*}
        \| PP^\dagger f' - f'\|_2 = O(NL_0/\sqrt{n}).
    \end{align*}
\end{lemma}

\begin{proof}
    This result follows from Lemma~\ref{lem:poly_approx_hippo}.
\end{proof}

Besides, it's easy to obtain an optimal function that satisfies $\E_{f \in {\cal D}}[ \| F^*(f_x) - f_y \|_2^2 ] \leq \epsilon$ where $\epsilon > 0$ is an arbitrary error. We especially focus on a kernel function as the following definition:
\begin{definition}\label{def:f_optimal}
    Denote $h > 0$ as the bandwidth of the kernel. We here define: $F^*(f_x) := (M({\cal I}_y) P) \cdot (P^\dagger \Phi(f_x))$, where $\Phi(f_x) := \frac{ \sum_{f_x' \in {\cal D}}{\cal K}(f_x, f_x')\cdot f }{\sum_{f_x' \in {\cal D}}{\cal K}(f_x, f_x')}$, and ${\cal K}(f_x, f_x') := \exp(-\frac{1}{2h} \|f_x - f_x'\|_2^2 )$ is a Gaussian kernel with kernel function.
\end{definition}
Thus, we could achieve a nearly-zero error $\epsilon > 0$ by choosing $h$ (bandwidth of kernel function) and $n$ (the order of polynomials $P$), such that:
\begin{lemma}\label{lem:minimum_func}
    Let function $F^*$ be defined as Definition~\ref{def:f_optimal}. Denote the failure probability $\delta \in (0, 0.1)$ and error $\epsilon > 0$. We choose $n = \Omega( \epsilon^{-1} N (L_0 + \sqrt{v\log(N/\delta)}/\Delta)  )$, where $v$ is the variance of noise under Assumption~\ref{asp:gaussian_noise} and $\Delta$ is the sample step size. Then we can show that with a probability at least $1 - \delta$, there exists a $h > 0$ satisfying $\E_{f \in {\cal D}}[\| F^*(f_x) - f_y \|_2^2] \leq \epsilon$ (Definition~\ref{def:problem}). 
\end{lemma}

\begin{proof}
    Holding the fact that with a probability at least $1 - \delta$,  the Gaussian tail bound:
    \begin{align*}
        |\xi_\tau| \leq O(\sqrt{v\log(N/\delta)}), \forall \tau \in [N].
    \end{align*}
    Hence, we have ($\forall \tau, \tau' \in [N]$):
    \begin{align*}
        |f_{\tau} - f_{\tau'}| \leq O(L_0 + \sqrt{v\log(N/\delta)} / \Delta) \cdot | \tau - \tau' |,
    \end{align*}
    where this step follows from the step size $\Delta > 0$.

    Finally, the result of this lemma will be achieved by plugging $n = \Omega( \epsilon^{-1} N (L_0+\sqrt{v\log(N/\delta)}/\Delta)  )$ and a small enough value of $h$ since some simple algebras.
\end{proof}
\section{Flow Matching for Time Series Generation}\label{sec:flow}

In this section, we introduce the core framework and methodology for time series generation using conditional flow with polynomial regularity, followed by the training objective and a sampling algorithm. In Section~\ref{sec:flow:cf_poly}, we define the conditional flow for time series generation, introducing the time-dependent mean and standard deviation functions, and the polynomial regularization of the flow. In Section~\ref{sec:flow:train_poly}, we specify the training objective based on the Flow Matching framework, defining the loss function and providing the closed-form solution for the optimal model. Finally, in Section~\ref{sec:flow:sampling}, we present the sampling algorithm for generating time series, utilizing the previously defined conditional flow and training objective.

\subsection{Conditional Flow with Polynomial Regularity}\label{sec:flow:cf_poly}

First, we define a important matrix $G$ as follows:
\begin{definition}\label{def:G}
    Let the output indices set ${\cal I}_y$ be defined in Definition~\ref{def:indices_set}. Let the polynomial basis $P$ be defined in Definition~\ref{def:poly_base}. Let the observation matrix $M({\cal I}) := [e_\tau^\top]_{\tau \in {\cal I}} \in \R^{|{\cal I}| \times N}$.
    We define the matrix $G \in \R^{N_y \times n}$ as 
    \begin{align*}
        G := M({\cal I}_y) P 
    \end{align*}
\end{definition}

Specially, we define the time-dependent mean of Gaussian distribution satisfying an ordinary equation. It is also called as our polynomial regularization.
\begin{definition}[Time-dependent mean of Gaussian distribution]\label{def:mean}
    Let $f = [f_x^\top, f_y^\top]^\top \in \R^{N}$ be defined as Definition~\ref{def:f}. Let $\alpha \in (0, 1)$ be some constant. Let $G$ be defined in Definition~\ref{def:G}.
    We define the time-dependent mean of Gaussian distribution as $\mu: [0, T] \times \R^{N} \rightarrow \R^{N_y}$, which satisfied the ODE that 
    \begin{align*}
        \mu_t'(f) = \alpha \cdot G G^\top ( GG^\dagger \psi_t(f) - f_y ),
    \end{align*}
\end{definition}

Meanwhile, we define the time-dependent standard deviation controls the uncertainty in the distribution, starting from a broad variance and gradually narrowing to a minimum value, which helps regulate the learning dynamics and stabilize the model.
\begin{definition}[Time-dependent standard deviation]\label{def:sd}
    Let $f = [f_x^\top, f_y^\top]^\top \in \R^{N}$ be defined as Definition~\ref{def:f}. Let $t \sim \mathsf{Uniform}[0,T]$. Let $\sigma_t: \R^{N} \rightarrow \R$.
    We define the minimum standard deviation $\sigma_{\min}$ as:
    \begin{align*}
        \sigma_{\min} := \sigma_1 (f).
    \end{align*}
      We define the time-dependent standard deviation $\sigma$ as
    \begin{align*}
        \sigma_t(f) := 1 - (1 - \sigma_{\min})t.
    \end{align*}
\end{definition}

Given data distribution with any time series data, $f \in {\cal D}$. The flow matching for time series generation~\cite{gtl24} defines a flow $\psi: [0, 1] \times \R^N$ taking time $t$ and time series data as input, matching $\psi_0(f) \sim \mathcal{N}(0, I_{N_y})$ at the beginning and $\psi_1(f) = f_y$ in the end, and then applying some neural networks to fit this distribution-to-distribution process. The detailed definition is given by:
\begin{definition}\label{def:flow}
    Let $f = [f_x^\top, f_y^\top]^\top \in \R^{N}$ be defined as Definition~\ref{def:f}. Let $\mu_t (f)$ be defined in Definition~\ref{def:mean}. Let $\sigma_t (f)$ be defined in Definition~\ref{def:sd}. Let $z \sim \N(0,I_{N_y})$ be the sample.
    We define the flow $\psi_t (f) \in \R^{N_y}$ as follows:
    \begin{align*}
        \psi_t (f) := \sigma_t (f) \cdot z + \mu_t (f).
    \end{align*}
\end{definition}

\subsection{Training Objective with Polynomial Regularity}\label{sec:flow:train_poly}

We slightly deviate from standard notation by defining the model function $F_\theta: \R^{N_y} \times \R^{N_x} \times [0, 1] \rightarrow \R^{N_y}$, parameterized by $\theta$, to capture the polynomial regularized conditional flow $\psi_t(f)$ introduced in Definition~\ref{def:flow}. This function takes the flow along with a temporal input to infer the corresponding vector field. The training procedure employs the Flow Matching framework~\cite{lcb+23}, which strives to shrink the discrepancy between the model’s estimates and the actual derivative of the flow.

Consequently, we define the training objective as the expected squared $\ell_2$ norm of the discrepancy:
\begin{definition}[Training Objective]\label{def:loss}
        Let $t \sim \mathsf{Uniform}[0,T]$.
        Let $f = [f_x^\top, f_y^\top]^\top \in \R^{N}$ be defined as Definition~\ref{def:f}.
        Let $z \sim \N(0,I_{N_y})$ be the sample.
        Let $\psi_t (f)$ be defined in Definition~\ref{def:flow}.
        Let $F_\theta: \R^{N_y} \times \R^{N_x} \times [0,T] \to \R^{N_y}$ be the model with parameter $\theta$.
    We define the training objective as follows:
    \begin{align*}
        &{\cal L}(\theta) :=  \E_{z, t, f}[\| F_\theta (\psi_t(f), f_x, t) - \frac{\d}{\d t} \psi_t(f) \|_2^2],
    \end{align*}
\end{definition}

We then provide the closed-form solution for $F_\theta$ that achieves the minimum of ${\cal L}(\theta)$ as follows:
\begin{theorem}[Informal version of Theorem~\ref{thm:formal:close_form_F}]\label{thm:close_form_F}
    Let ${\cal L}(\theta)$ be defined in Definition~\ref{def:loss}. 
    Let $z \sim \mathcal{N}(0, I_{N_y})$. Let $t \sim \mathsf{Uniform}[0,T]$. Let $f_x,~f_y$ be defined in Definition~\ref{def:f}. Let $G$ be defined in Definition~\ref{def:G}. Let $\sigma_{\min}$ be defined in Definition~\ref{def:sd}. The optimal $F_\theta$ that minimizes ${\cal L}(\theta)$  satisfies: 
    \begin{align*}
        F_\theta(z, f_x, t) = GG^\top\!\bigl( GG^\dagger z - f_y \bigr) + (\sigma_{\min} - 1)\, z.
    \end{align*}
\end{theorem}

\subsection{Sampling Algorithm}\label{sec:flow:sampling}

Now we review the algorithm form of the sampling process of flow matching for time series generation in Algorithm~\ref{alg:sampling}.

\begin{algorithm}[!ht]\caption{Recall the sampling process of flow matching for time series generation}\label{alg:sampling}

\begin{algorithmic}[1]
\Statex \textbf{Input: } Time series $f_x \in \R^{N_x}$, sample steps $T > 0$

\Statex \textbf{Output: } Predictive time series $x_1 \in \R^{N_y}$

\Procedure{Sampling}{$f_x$}

\State Sample the initial Gaussian noise $x_0 \in \mathcal{N}(0, I_{N_y})$

\For{$t \in [T]$}

\State If $t > 1$, sample $z \sim \mathcal{N}(0, I_y)$; else, $z = {\bf 0}_{\bf N_y}$

\State Update $x_{\frac{t}{T}} \leftarrow x_{\frac{t-1}{T}} - T \cdot F_\theta(x_{\frac{t-1}{T}}, f_x, \frac{t-1}{T})$

\State Update $x_{\frac{t}{T}} \leftarrow x_{\frac{t}{T}} - (1 - (1 - \sigma_{\min})\frac{t}{T}) \cdot z$

\EndFor

\State \Return $x_1$

\EndProcedure

\end{algorithmic}
\end{algorithm}

\section{Approximation}\label{sec:app}

In this section, we utilize the approximation ability of the transformer-based neural networks, especially, Diffusion Transformer (DiT). First, in Section~\ref{sec:appro:dit}, we present the DiT backbone, a widely adopted model in empirical research. Next, we introduce the main theorem in Section~\ref{sec:appro:main}, which provides an approximation result and establishes an upper bound on the error.

\subsection{Diffusion Transformer (DiT)}\label{sec:appro:dit}

Diffusion Transformer~\cite{px23} introduces a strategy where Transformers~\cite{vsp+17} serve as the core architecture for Diffusion Models~\cite{ddpm,ddim}. In particular, each Transformer block comprises a multi-head self-attention module and a feed-forward component, both of which include skip connections. We first define the multi-head self-attention:
\begin{definition}[Multi-head self-attention]\label{def:attn}
    Given $h$-heads query, key, value and output projection weights $W_Q^i, W_K^i, W_V^i, W_O^i  \in \R^{d \times m}$ with each weight is a $d \times m$ shape matrix, for an input matrix $X \in \R^{L \times d}$, we define a multi-head self-attention $\mathsf{Attn} : \R^{L \times d} \to \R^{L \times d}$ as follows:
    \begin{align*}
        {\sf Attn}(X) :=  \sum_{i=1}^h {\sf Softmax}( X W_Q^i {W_K^i}^\top X^\top ) \cdot X W_V^i {W_O^i}^\top + X.
    \end{align*}
\end{definition}
A feed-forward layer transforms input data by applying linear projections, a non-linear activation function, and residual connections, which is defined as follows:
\begin{definition}[Feed-forward]\label{def:feed_forward}
    Given two projection weights $W_1, W_2 \in \R^{d \times r}$ and two bias vectors $b_1 \in \R^r$ and $b_2 \in \R^{d}$, for an input matrix $X \in \R^{L \times d}$, we define a feed-forward computation ${\sf FF} : \R^{L \times d} \to \R^{L \times d}$ follows:
    \begin{align*}
        {\sf FF}(X) := \phi(X W_1 + {\bf 1}_L b_1^\top) \cdot W_2^\top + {\bf 1}_L b_2^\top + X.
    \end{align*}
    Here, $\phi$ is an activation function and usually be considered as ReLU.
\end{definition}

We denote a Transformer block as ${\sf TF}^{h, m, r}: \R^{L \times d}\rightarrow \R^{L \times d}$, where $h$ is the count of attention heads, $m$ specifies the head dimension within the self-attention mechanism, and $r$ is the hidden size in the feed-forward layer. Building on multi-head self-attention and the feed-forward layer, we define the transformer block as follows:
\begin{definition}[Transformer block]\label{def:transformer_tf}
    Let multi-head self-attention and feed-forward neural network be defined in Definition~\ref{def:attn} and Definition~\ref{def:feed_forward} respectively. Formally, for an input matrix $X \in \R^{L \times d}$, we define the Transformer block ${\sf TF}^{h, m, r} : \R^{L \times d} \to \R^{L \times d}$:
    \begin{align*}
        {\sf TF}^{h, m, r}(X) := {\sf FF} \circ {\sf Attn}(X)
    \end{align*}
\end{definition}

 We define the Transformer network as the composition of Transformer blocks:
\begin{definition}[Complete transformer network]\label{def:model}
    We consider a transformer network as a composition of a transformer block (Definition~\ref{def:transformer_tf}) with model weight $\theta^{h, m, r}$, which is:
    \begin{align*}
        {\cal T}^{h, m, r} := \{ {\cal F}: \R^{L \times d} \rightarrow \R^{L \times d}~|~ &\text{${\cal F}$ is a composition of Transformer blocks ${\sf TF}_{\theta^{h, m, r}}$’s}\\
        ~ & \text{with positional embedding $E \in \R^{L \times d}$}\}
    \end{align*}
\end{definition}
In this paper, Transformer networks with positional encoding $E\in\R^{L \times d}$ is used in the analysis.
We take a Transformer network consisting $K$ blocks and positional encoding as an example:
\begin{example}\label{exp:cal_F}
    We here give an example for the sequence-to-sequence mapping $f_{{\cal T}}$ in Definition~\ref{def:model}: Denote $K$ as the number of layers in some transformer network. For an input matrix $X \in \R^{L \times d}$, we use $E \in \R^{L \times d}$ to denote the positional encoding, we then define:
    \begin{align*}
      f_{{\cal T}}(X)= {\sf TF}^{h, m, r}_{(K)} \circ  \cdots \circ {\sf TF}^{h, m, r}_{(1)} (X+E).
    \end{align*}
\end{example}

\subsection{Main Theorem I: Approximation}\label{sec:appro:main}

We first present the universal approximation theorem for transformer-based models and utilize it as a lemma to establish our main theorem..
\begin{lemma}[Theorem 2 of \cite{ybr+20}]\label{lem:uat}
    Let $\epsilon > 0$ and let ${\cal F}_\mathrm{PE}$ be the function class consisting all continuous permutation equivariant functions with compact support that $\R^{L \times d} \to \R^{L \times d}$. For any $f,g : \R^{L \times d} \to \R^{L \times d}$ be two different functions, we can show that for any given $f \in {\cal F}_\mathrm{PE}$, there exists a Transformer $g \in {\cal T}^{h,m,r}$ such that
    \begin{align*}
        \|f(X) - g(X) \|_2 \leq \epsilon, \forall X \in \R^{L \times d}.
    \end{align*}
\end{lemma}

Before we state the approximation theorem, we define a reshaped layer that transforms concatenated input in flow matching into a length-fixed sequence of vectors. 
\begin{definition}[DiT reshape layer]\label{def:reshape}
    Let $R: \R^{N+1} \rightarrow \R^{n \times d}$ be a reshape layer that transforms the $(N+1)$-dimensional input vector into a $n \times d$ matrix. 
\end{definition}

Therefore, in the following, we give the theorem utilizing DiT to minimize training objective ${\cal L}(\theta)$ to arbitrary error.

\begin{theorem}[Informal version of Theorem~\ref{thm:formal:main_1}]\label{thm:main_1}
    Let all pre-conditions hold in Lemma~\ref{lem:minimum_func}. Let the DiT reshape layer $R$ be defined in Definition~\ref{def:reshape}. There exists a transformer network $f_{\cal T} \in {\cal T}_{P}^{2, 1, 4}$ defining function $F_\theta(z, f_x, t) := f_{\cal T}( R([z^\top, f_x^\top, t]^\top) )$ with parameters $\theta$ that satisfies ${\cal L}(\theta) \leq \epsilon$ for any error $\epsilon > 0$. 
\end{theorem}

\section{Generalization}\label{sec:gen}

This section establishes generalization guarantees for the transformer-based sampling algorithm by combining analytical tools and convergence results. Section~\ref{sec:gen:tool} introduces foundational bounds on pseudoinverse matrices and derives an error bound $\epsilon_1$ for the regularized function $\wh{F}$ under noisy sampling, while Section~\ref{sec:gen:gen} leverages these bounds to prove the transformer network’s asymptotic generalization error $\epsilon_0 + \epsilon_1$, connecting algorithmic stability with approximation-theoretic guarantees.

\subsection{Basic Tools}\label{sec:gen:tool}
We now provide two lemmas as the toolkit for proving the generalization.
\begin{lemma}[Informal version of Lemma~\ref{fac:formal:infity_norm_pesdueo_inverse}]\label{fac:infity_norm_pesdueo_inverse}
    For a PD matrix $A \in \R^{d_1 \times d_2}$ with a positive minimum singular value $\lambda_{\min}(A) > 0$, the infinite norm of its pseudoinverse matrix $A^\dag$ is given by:
    \begin{align*}
        \| A^\dagger \|_\infty \leq \frac{1}{\lambda_{\min}(A)}.
    \end{align*}
\end{lemma}

\begin{lemma}[Informal version of Lemma~\ref{fac:formal:pesdueo_inverse_diff}]\label{fac:pesdueo_inverse_diff}
    For two matrices $A , B \in\R^{d_1 \times d_2}$, we have:
    \begin{align*}
        \| A^\dagger - B^\dagger \| \leq \max\{ \| A^\dagger \|^2, \| B^\dagger \|^2  \}\cdot \| A - B \|.
    \end{align*}
\end{lemma}

Thus, we define another regularized function $\hat{F}(f_x) := M({\cal I}_y) P (M({\cal I}_x) P)^\dagger f_x$, then we have:
\begin{lemma}[Informal version of Lemma~\ref{lem:formal:generalize_hat_F}]\label{lem:generalize_hat_F}
    Let $\delta \in (0,0.1)$. For any in-distribution (ID) data $f \in {\cal D}$ be defined in Definition~\ref{def:f} and its corresponding signal $g: \R_{\ge 0} \rightarrow \R$, we sample new data $\wt{f} := [g(\tau \cdot \Delta) + \xi_\tau]$, we first define:
    \begin{align*}
        \epsilon_1 := O\Big(\frac{\sqrt{N_yv\log(N/\delta)}\exp(O(nN))}{\lambda} + \frac{N^{1.5}L}{\sqrt{n}}\Big)^2.
    \end{align*}
    where $v$ is the variance of noise under Assumption~\ref{asp:gaussian_noise} and $\Delta$ is the sample step size. Then with a probability at least $1 - \delta$, we have
    \begin{align*}
        \E_{f \in {\cal D}}[\| \hat{F}(\wt{f}_x) - \wt{f}_y \|_2^2] 
        \leq \epsilon_1.
    \end{align*}
\end{lemma}

\subsection{Main Theorem II: Generalization}\label{sec:gen:gen}
We present our generalization result as follows:
\begin{theorem}\label{thm:main_2}
    Denote the failure probability $\delta \in (0, 0.1)$ and an arbitrary error $\epsilon_0 > 0$. There exists a transformer network $f_{\cal T} \in {\cal T}_{P}^{2, 1, 4}$ defining function $F_\theta(z, f_x, t) := f_{\cal T}( R([z^\top, f_x^\top, t]^\top) )$ with parameters $\theta$ that satisfies:  for any in-distribution (ID) data $f \in {\cal D}$ and its corresponding signal $g: \R_{\ge 0} \rightarrow \R$, we sample new data $\wt{f} := [g(\tau \cdot \Delta) + \xi_\tau]$, where $\Delta$ is the sample step size. We denote $x_1$ as the output of Algorithm~\ref{alg:sampling} with $T$ steps. Then with a probability at least $1 - \delta$, we have:
        \begin{align*}
            \lim_{T \rightarrow +\infty} \E_{x_0 \sim \mathcal{N}(0, I_{N_y}), f\in {\cal D}}[\| x_1 - \wt{f}_y\|_2^2] \leq \epsilon_0 + \epsilon_1
        \end{align*}
\end{theorem}

\begin{proof}
    This proof combines from Lemma~\ref{lem:generalize_hat_F} and other proofs are similar with the ones in Theorem~\ref{thm:main_1} since we suggest the transformer network to represent the function $\hat{F}$.
\end{proof}

\section{Efficiency}\label{sec:eff}

Here in this section, we consider the sampling efficiency problem of the vanilla sampling process of flow matching for time series generation (Algorithm~\ref{alg:sampling}).

This section analyzes the convergence properties of the sampling algorithm through gradient descent, establishing error decrease and overall efficiency. Section~\ref{sec:eff:error} analyzes the error decrease per iteration by establishing gradient descent updates and key properties including Lipschitz smoothness, unbiased updates, and update norms, while Section~\ref{sec:eff:convergence} establishes the overall convergence rate of the algorithm by bounding the minimum expected gradient norm across iterations, demonstrating efficiency under chosen parameters.

\subsection{Error Decrease}\label{sec:eff:error}

{\bf Gradient descent with respect to some objective. }
As we define the  polynomial regularization in Definition~\ref{def:mean}, we claim that Algorithm~\ref{alg:sampling} implements a first-order gradient descent to some implicit parameter, we denote it as $w: [T] \rightarrow \R^n$. Formally, we define $w$ as 
\begin{definition}[Implicit parameter $w$]\label{def:w}
    Let $P$ be defined in Definition~\ref{def:poly_base}. Let $f_y$ be defined in Definition~\ref{def:f}. We denote the implicit parameter $w$ as $w: [T] \rightarrow \R^n$, i.e., $w_t \in \R^n$ for time step $t$. Particularly, we define $w_0 := P^\dagger x_0$ as the initialization and $w^* := P^\dagger f_y$ as the optimal solution.
\end{definition}

Besides, we use the metric that measures the square $\ell_2$ norm of the difference between the current sampling result $x_{\frac{t}{T}}$ and the ground truth. Formally, we define the metric as follows:
\begin{definition}[Metric]\label{def:metric}
    Let $w$ be defined in Definition~\ref{def:w}. Let $P$ be defined in Definition~\ref{def:poly_base}. Let $f$ and $f_y$ be defined in Definition~\ref{def:f}. We define the metric $u: \R^n \to \R$ as 
    \begin{align*}
        u(w_t) := \E_{f \in {\cal D}}[\| P w_t - f_y \|_2^2].
    \end{align*}
\end{definition}

Then the update is given by:
\begin{definition}[Update Rule]\label{def:update}
Let $w$ be defined in Definition~\ref{def:w}. Let $P$ be defined in Definition~\ref{def:poly_base}. Let $F_\theta: \R^{N_y} \times \R^{N_x} \times [0,T] \to \R^{N_y}$ be the model with parameter $\theta$. Let $\sigma_t$ be the time-dependent standard deviation. Let $f_x$ and $f_y$ be defined in Definition~\ref{def:f}. Let $z \sim \N(0,I_{N_y})$ be the sample.
We use $\Delta w_t$ to denote the weight adjustment, which is defined as
\begin{align*}
    \Delta w_{t-1} := P^\dagger \Big(T \cdot F_\theta (Pw_{t-1}, f_x, \frac{t-1}{T})  + z \cdot \sigma_{\frac{t}{T}}(f) \Big).
\end{align*}
In each iteration, we update the parameter as
\begin{align*}
    w_t =  w_{t-1} - \Delta w_{t-1}.
\end{align*}
\end{definition}

\begin{lemma}
    Let $w$ be defined in Definition~\ref{def:w}. Let $\alpha$ be the constant in Definition~\ref{def:mean}. Let $P$ be defined in Definition~\ref{def:poly_base}. Let $F_\theta: \R^{N_y} \times \R^{N_x} \times [0,T] \to \R^{N_y}$ be the model with parameter $\theta$. Let $f_x$ and $f_y$ be defined in Definition~\ref{def:f}. Let $G$ be defined in Definition~\ref{def:G}. We can show that
    \begin{align*}
        \| P^\dagger F_\theta (Pw_{t}, f_x, \frac{t}{T}) - \alpha G^\top(G w_{t} - f_y) \|_2^2\leq \epsilon_0,
    \end{align*}
    where $\epsilon_0 > 0$ is an arbitrary positive error.
\end{lemma}
\begin{proof}
    This result follows from Lemma~\ref{lem:uat}.
\end{proof}

First, we give the some tools in helping the analysis as follows:

\begin{lemma}[Informal version of Lemma~\ref{lem:formal:tool}]\label{lem:tool}
    Let $w$ be defined in Definition~\ref{def:w}. Let $t,t' \in [0,T]$ be two different time step. Let $u(w_t)$ be defined in Definition~\ref{def:metric}. Let $\lambda := \lambda_{\min}(P) > 0$. Let $\alpha$ be the constant in Definition~\ref{def:mean}. Let $G$ be defined in Definition~\ref{def:G}. Let $\sigma_t$ be defined in Definition~\ref{def:sd}. Let $\Delta w_t$ be defined in Definition~\ref{def:update}. Let $f$ be defined in Definition~\ref{def:f}.
    Then we have
    \begin{itemize}
        \item {\bf Lipschitz-smooth. } $\forall w_{t}, w_{t'} \in \R^n$,
        \begin{align*}
             \| \nabla_{w_{t}} u(w_{t}) - \nabla_{w_{t'}} u(w_{t'}) \|_2 \leq \frac{n\exp(O(nN))}{\lambda} \|w_{t} - w_{t'}\|_2.
        \end{align*}
        \item {\bf Unbiased update. } 
        \begin{align*}
            \E[\Delta w_t] = \alpha T \cdot \E[\nabla_{w_t} u(w_t)].
        \end{align*}
        \item {\bf Update norm. } 
        \begin{align*}
            \E[\|\Delta w_t\|_2^2] = \alpha^2 T^2 \cdot \E[\| \nabla_{w_t} u(w_t)\|_2^2] + n \cdot \sigma_{\frac{t}{T}}(f).
        \end{align*}
    \end{itemize}
\end{lemma}

Thus, we prove the expectation of error decrease of sampling at each step, as we state below:
\begin{lemma}[Informal version of Lemma~\ref{lem:formal:decrease}]\label{lem:decrease}
    We define $L_1 := n \cdot \frac{\exp(O(nN))}{\lambda}$. Let $w$ be defined in Definition~\ref{def:w}. Let $u(w_t)$ be defined in Definition~\ref{def:metric}.  Let $\alpha$ be the constant in Definition~\ref{def:mean}. Let $\sigma_t (f)$ be defined in Definition~\ref{def:sd}. Let $f$ be defined in Definition~\ref{def:f}.
    Let all pre-conditions hold in Lemma~\ref{lem:minimum_func}. For each step $t \in [T]$, we have:
    \begin{align*}
        \E[u(w_t)] \leq \E[u(w_{t-1})] + (\frac{L_1}{2}  \alpha^2 T^2- \alpha T) \E[\|\nabla_{w_{t-1}} u(w_{t-1})\|_2^2] + \frac{L_1 n}{2}  \sigma_{\frac{t-1}{T}}(f)
    \end{align*}
\end{lemma}

\subsection{Main Theorem III: Convergence}\label{sec:eff:convergence}

Here, we demonstrate the efficiency of the sample process below:
\begin{theorem}[Informal version of Theorem~\ref{thm:formal:main_3}]\label{thm:main_3}
    Let $w$ be defined in Definition~\ref{def:w}. Let $u(w_t)$ be defined in Definition~\ref{def:metric}.
    Let $\delta \in (0,0.1)$. Let all pre-conditions hold in Lemma~\ref{lem:minimum_func}. Denote the failure probability $1 - \delta$. For error $\epsilon > 0$, we choose $T = \wt{O}(\sqrt{N/(L_1\alpha \epsilon)})$, then with a probability at least $1 - \delta$, we have:
    \begin{align*}
        \min_{t \in [T]}\E[\|\nabla_{w_t} u(w_t)\|_2^2] \leq \epsilon
    \end{align*}
\end{theorem}

\section{Conclusion}\label{sec:conclusion}
This paper establishes a theoretical framework for understanding flow-based generative models in time series analysis, addressing the critical gap between empirical success and theoretical foundations. By integrating polynomial regularization into the flow matching objective, we demonstrate that transformer-based architectures can achieve provable approximation, generalization, and convergence guarantees. Our analysis reveals three key insights: (1) Diffusion Transformers universally approximate the optimal flow matching objective, (2) polynomial regularization enables generalization bounds combining approximation errors and noise tolerance, and (3) the sampling process exhibits gradient descent-like convergence under Lipschitz smoothness conditions. These results provide the first end-to-end theoretical justification for modern time series generation paradigms, confirming that architectural choices like DiT and training strategies like flow matching jointly enable both expressivity and stability. Future work could extend this framework to non-Gaussian noise settings and investigate the tightness of our polynomial-dependent error bounds. More broadly, our methodology opens new avenues for theoretically grounding other temporal generative models while maintaining alignment with practical implementations.

\ifdefined\isarxiv
\bibliographystyle{alpha}
\bibliography{ref}
\else
\bibliography{ref}
\bibliographystyle{plainnat}

\fi

\newpage
\onecolumn
\appendix

\ifdefined\isarxiv

\section*{Appendix}
\else

\title{Theoretical Foundation of Flow-Based Time Series Generation: Provable Approximation, Generalization, and Efficiency\\(Supplementary Material)}
\maketitle

\fi

\paragraph{Roadmap.}Section~\ref{sec:app:calculation} present two useful norm facts. Section~\ref{sec:app:close_form} present the optimal solution of the neural network. Section~\ref{sec:app:missing_proof} present the missing proof of our main results.

\section{Basic Calculation}\label{sec:app:calculation}
\begin{lemma}[Formal version of Lemma~\ref{fac:infity_norm_pesdueo_inverse}]\label{fac:formal:infity_norm_pesdueo_inverse}
    For a PD matrix $A \in \R^{d_1 \times d_2}$ with a positive minimum singular value $\lambda_{\min}(A) > 0$, the infinite norm of its pseudoinverse matrix $A^\dag$ is given by:
    \begin{align*}
        \| A^\dagger \| \leq \frac{1}{\lambda_{\min}(A)}.
    \end{align*}
\end{lemma}

\begin{proof}
    We have:
    \begin{align*}
        \| A^\dagger \|
        = & ~ \| (U \Sigma V)^\dagger \| \\
        = & ~ \| V^\top \Sigma^\dagger U^\top \| \\
        = & ~ \| \Sigma^\dagger \| \\
        \leq & ~ \frac{1}{\lambda_{\min}(A)}
    \end{align*}
    where the first step follows from the svd of $A = U \Sigma V$, the second step follows from simple algebras, the third step follows from $U, V$ are orthogonal (and square) matrices, the last step follows from the definitions of the spectral norm and $\Sigma$ is a diagonal matrix of singular values.
\end{proof}

\begin{lemma}[Formal version of Lemma~\ref{fac:pesdueo_inverse_diff}]\label{fac:formal:pesdueo_inverse_diff}
    For two matrices $A , B \in\R^{d_1 \times d_2}$, we have:
    \begin{align*}
        \| A^\dagger - B^\dagger \| \leq \max\{ \| A^\dagger \|^2, \| B^\dagger \|^2  \}\cdot \| A - B \|.
    \end{align*}
\end{lemma}

\begin{proof}
    We have:
    \begin{align*}
        \| A^\dagger - B^\dagger \| 
        \leq & ~ \| A^\dagger \| \cdot \| I_{d_1} - A B^\dagger \| \\
        \leq & ~ \| A^\dagger \| \| B^\dagger \| \cdot \| A - B \| \\
        \leq & ~ \max\{ \| A^\dagger \|^2, \| B^\dagger \|^2  \}\cdot \| A - B \|
    \end{align*}
    where these steps follow from simple algebras and $A^\dagger A \approx I_{d_1}$
\end{proof}

\section{Close Form of Optimal Solution}\label{sec:app:close_form}

We then provide the closed-form solution for $F_\theta$ that achieves the minimum of ${\cal L}(\theta)$ as follows:
\begin{theorem}[Formal version of Theorem~\ref{thm:close_form_F}]\label{thm:formal:close_form_F}
If the following conditions hold:
\begin{itemize}
    \item Let ${\cal L}(\theta)$ be defined in Definition~\ref{def:loss}.
    \item Let $z \sim \mathcal{N}(0, I_{N_y})$.
    \item Let $t \sim \mathsf{Uniform}[0,T]$.
    \item Let $G$ be defined in Definition~\ref{def:G}.
    \item Let $f_x, f_y$ be defined in Definition~\ref{def:f}.
    \item Let $\sigma_{\min}$ be defined in Definition~\ref{def:sd}.
\end{itemize}
   The optimal $F_\theta$ that minimizes ${\cal L}(\theta)$  satisfies:
    \begin{align*}
        F_\theta(z, f_x, t) = GG^\top\!\bigl( GG^\dagger z - f_y \bigr) + (\sigma_{\min} - 1)\, z.
    \end{align*}
\end{theorem}
\begin{proof}
    Observe that
    \begin{align*}
        \psi'_t(f) 
        = & ~ \mu'_t(f) + \sigma_t'(f) \cdot z \\
        = & ~ GG^\top ( GG^\dagger \psi_t(f) - f_y ) + (\sigma_{\min} - 1) z,
    \end{align*}
    where the initial step follows from the construction and definition of $\psi_t(f)$, and the subsequent step is due to Definition~\ref{def:mean}. Substituting $\psi_t(f)$ with $z$ completes the derivation.
\end{proof}

\section{Missing Proofs}\label{sec:app:missing_proof}
In Section~\ref{sec:app:miss:appro}, we present the missing proof in Section~\ref{sec:app}. In Section~\ref{sec:app:miss:gen}, we present the missing proof in Section~\ref{sec:gen}. In Section~\ref{sec:app:miss:eff}, we present the missing proof in Section~\ref{sec:eff}. 

\subsection{Approximation}\label{sec:app:miss:appro}
\begin{theorem}[Formal version of Theorem~\ref{thm:main_1}]\label{thm:formal:main_1}
    If the following conditions hold:
    \begin{itemize}
        \item Let all pre-conditions hold in Claim~\ref{lem:minimum_func}.
        \item Let the DiT reshape layer $R$ be defined in Definition~\ref{def:reshape}.
    \end{itemize}
      Then there exists a transformer network $f_{\cal T} \in {\cal T}_{P}^{2, 1, 4}$ defining function $F_\theta(z, f_x, t) := f_{\cal T}( R([z^\top, f_x^\top, t]^\top) )$ with parameters $\theta$ that satisfies ${\cal L}(\theta) \leq \epsilon$ for any error $\epsilon > 0$. 
\end{theorem}

\begin{proof}
    Choose $L=1$ for $R(\cdot)$, we define:
    \begin{align*}
        f_{\cal T}^*([z^\top, f_x^\top, t]^\top)
        := & ~  GG^\top ( GG^\dagger z - F^*(f_x) ) + (\sigma_{\min} - 1) z.
    \end{align*}
    Then, following Lemma~\ref{lem:uat}, there exists a transformer network $f_{\cal T} \in {\cal T}_{P}^{2, 1, 4}$ that satisfies (arbitrary error $\epsilon > 0$):
    \begin{align*}
        \| f_{\cal T}( R([z^\top, f_x^\top, t]^\top) ) - f_{\cal T}^*([z^\top, f_x^\top, t]^\top)\|_2 \leq \epsilon.
    \end{align*}
    Since $\|P\|_\infty \leq \exp(O(nN))$, we have $\|GG^\top\|_2 \leq N \exp(O(nN))$, scaling $\epsilon \leq \frac{\epsilon_0}{N \exp(O(nN))}$ could directly achieve the theorem result.
\end{proof}

\subsection{Generalization}\label{sec:app:miss:gen}

\begin{lemma}[Formal version of Lemma~\ref{lem:generalize_hat_F}]\label{lem:formal:generalize_hat_F}
    If the following conditions hold:
    \begin{itemize}
        \item Let $\delta \in (0,0.1)$.
        \item Let $\epsilon_1 := O\Big(\frac{\sqrt{N_yv\log(N/\delta)}\exp(O(nN))}{\lambda} + \frac{N^{1.5}L}{\sqrt{n}}\Big)^2$ be the error bound, where $v$ is the variance of noise under Assumption~\ref{asp:gaussian_noise}.
        \item Let in-distribution (ID) data $f \in {\cal D}$ be defined in Definition~\ref{def:f}.
        \item Let $g: \R_{\ge 0} \rightarrow \R$ be the corresponding signal of $f$.
        \item Let $\wt{f} := [g(\tau \cdot \Delta) + \xi_\tau]$ be a new sampled data, where $\Delta$ is the sample step size.
    \end{itemize}
    Then with a probability at least $1 - \delta$, we have
    \begin{align*}
        \E_{f \in {\cal D}}[\| \hat{F}(\wt{f}_x) - \wt{f}_y \|_2^2] 
        \leq \epsilon_1.
    \end{align*}
\end{lemma}

\begin{proof}
    We have:
    \begin{align*}
        \E_{f \in {\cal D}}[\| \hat{F}(\wt{f}_x) - \wt{f}_y \|_2]
        \leq & ~ \E_{f \in {\cal D}}[\| M({\cal I}_y) P (M({\cal I}_x) P)^\dagger \wt{f}_x - M({\cal I}_y) P P^\dagger \wt{f} \|_2] + O(N^{1.5}L/\sqrt{n}) \\
        \leq & ~ \|P\| \cdot \E_{f \in {\cal D}}[\| (M({\cal I}_x) P)^\dagger \wt{f}_x - P^\dagger \wt{f} \|_2]+ O(N^{1.5}L/\sqrt{n}) \\
        \leq & ~ \|P\| \cdot \E_{f \in {\cal D}}[\| (M({\cal I}_x) P)^\dagger - P^\dagger \| \cdot \| \wt{f}_x\|_2 + \|P^\dagger \| \| M({\cal I}_x)^\dagger \wt{f}_x - \wt{f} \|_2] + O(N^{1.5}L/\sqrt{n})\\
        \leq & ~ \frac{\sqrt{N_yv\log(N/\delta)}\exp(O(nN))}{\lambda} + O(N^{1.5}L/\sqrt{n})
    \end{align*}
    where the first step follows from the polynomial approximation (Lemma~\ref{lem:poly_approx}), the second step follows from Cauchy-Schwarz inequality, the third step follows from simple algebras and triangle inequality, and the last step follows from some simple calculations with Lemma~\ref{fac:infity_norm_pesdueo_inverse} and Lemma~\ref{fac:pesdueo_inverse_diff}.
\end{proof}

\subsection{Efficiency}\label{sec:app:miss:eff}
\begin{lemma}[Formal version of Lemma~\ref{lem:tool}]\label{lem:formal:tool}
    If the following conditions hold:
    \begin{itemize}
        \item Let $w$ be defined in Definition~\ref{def:w}.
        \item Let $t,t' \in [0,T]$ be two different time step.
        \item Let $u(w_t)$ be defined in Definition~\ref{def:metric}.
        \item Let $\lambda := \lambda_{\min}(P) > 0$.
        \item Let $\alpha$ be the constant in Definition~\ref{def:mean}.
        \item Let $G$ be defined in Definition~\ref{def:G}.
        \item Let $\sigma_t$ be defined in Definition~\ref{def:sd}.
        \item Let $\Delta w_t$ be defined in Definition~\ref{def:update}.
        \item Let $f$ be defined in Definition~\ref{def:f}.
    \end{itemize}
    Then we have:
    \begin{itemize}
        \item {\bf Lipschitz-smooth. } $\forall w_{t}, w_{t'} \in \R^n$,
        \begin{align*}
             \| \nabla_{w_{t}} u(w_{t}) - \nabla_{w_{t'}} u(w_{t'}) \|_2 \leq \frac{n\exp(O(nN))}{\lambda} \|w_{t} - w_{t'}\|_2.
        \end{align*}
        \item {\bf Unbiased update. } 
        \begin{align*}
            \E[\Delta w_t] = \alpha T \cdot \E[\nabla_{w_t} u(w_t)].
        \end{align*}
        \item {\bf Update norm. } 
        \begin{align*}
            \E[\|\Delta w_t\|_2^2] = \alpha^2 T^2 \cdot \E[\| \nabla_{w_t} u(w_t)\|_2^2] + n \cdot \sigma_{\frac{t}{T}}(f).
        \end{align*}
    \end{itemize}
\end{lemma}

\begin{proof}
    {\bf Proof of gradient Lipschitz-smooth. } We have:
    \begin{align*}
        \| \nabla_{w_{t}} u(w_{t}) - \nabla_{w_{t'}} u(w_{t'}) \|_2 = & ~ \| G^\top (Gw_{t} - \E[f_y]) - G^\top(Gw_{t'} - \E[f_y]) \|_2 \\
        = & ~ \| G^\top (Gw_{t} - Gw_{t'}) \|_2 \\
        \leq & ~ \| G^\top G\|_2 \cdot \|w_{t} - w_{t'}\|_2 \\
        \leq & ~ \frac{n\exp(O(nN))}{\lambda} \|w_{t} - w_{t'}\|_2,
    \end{align*}
    where the first step follows from the derivation of $u(w)$, the second step follows from simple algebras, the third step follows from Cauchy-Schwarz inequality, the last step follows from $\| G\|_\infty \leq \exp(O(nN))$.

    {\bf Proof of unbiased update.} We have:
    \begin{align*}
        \E[\Delta w_t] = & ~ \E[P^\dagger \Big(T F(Pw_{t-1}, f_x, \frac{t-1}{T})  + \sigma_{\frac{t}{T}}(f) z\Big)] \\
        = & ~ \alpha T G^\top(G w_t - \E[f_y]) \\
        = & ~\alpha T \nabla_{w_t} u(w_t),
    \end{align*}
    where the first step follows from Definition~\ref{def:update}, the second step follows from $\E[z] = {\bf 0}_d$, the last step follows from the derivation of $u(w)$.

    {\bf Proof of update norm.} We have:
    \begin{align*}
        \E[\|\Delta w_t\|_2^2] = & ~ \alpha^2 T^2 \E[\| \nabla_{w_t} u(w_t)\|_2^2 ] - \alpha T\E[\sigma_{\frac{t}{T}}(f) \langle \nabla_{w_t} u(w_t), z \rangle] + \E[\| \sigma_{\frac{t}{T}}(f) z\|_2^2] \\
        = & ~ \alpha^2 T^2 \E[\| \nabla_{w_t} u(w_t)\|_2^2 ]  + \E[\| \sigma_{\frac{t}{T}}(f) z\|_2^2] \\
        = & ~ \alpha^2 T^2 \E[\| \nabla_{w_t} u(w_t)\|_2^2] + \sigma_{\frac{t}{T}}(f)n
    \end{align*}
    where the first step follows from Definition~\ref{def:update}, the second step follows from $\E[z] = {\bf 0}_d$, the last step follows from $E[\|z\|_2^2] = n$ (the variance of Gaussian distribution).
\end{proof}

\begin{lemma}[Formal version of Lemma~\ref{lem:decrease}]\label{lem:formal:decrease}
    If the following conditions hold:
    \begin{itemize}
        \item We define $L_1 := n \cdot \frac{\exp(O(nN))}{\lambda}$.
        \item Let $w$ be defined in Definition~\ref{def:w}.
        \item Let $u(w_t)$ be defined in Definition~\ref{def:metric}.
        \item Let $\alpha$ be the constant in Definition~\ref{def:mean}.
        \item Let $\sigma_t (f)$ be defined in Definition~\ref{def:sd}.
        \item Let $f$ be defined in Definition~\ref{def:f}.
        \item Let all pre-conditions hold in Lemma~\ref{lem:minimum_func}.
    \end{itemize}
    Then for each step $t \in [T]$, we have:
    \begin{align*}
        \E[u(w_t)] 
        \leq & ~ \E[u(w_{t-1})] + (\frac{L_1}{2}  \alpha^2 T^2- \alpha T) \E[\|\nabla_{w_{t-1}} u(w_{t-1})\|_2^2] + \frac{L_1 n}{2}  \sigma_{\frac{t-1}{T}}(f)
    \end{align*}
\end{lemma}

\begin{proof}
    We first give a common tool for proving convergence that is derived from Taylor expansion, such that:
    \begin{align*}
        u(w_t) \leq & ~ u(w_{t-1}) - \langle \nabla_{w_{t-1}} u(w_{t-1}), \Delta w_{t-1} \rangle + \frac{L_1}{2} \| \Delta w_{t-1}\|_2^2.
    \end{align*}

    Next, taking expectation to the whole equation, we can get:
    \begin{align*}
        \E[u(w_t)] \leq & ~ \E[u(w_{t-1}) - \langle \nabla_{w_{t-1}} u(w_{t-1}), \Delta w_{t-1} \rangle + \frac{L_1}{2} \| \Delta w_{t-1}\|_2^2]\\
        = & ~ \E[u(w_{t-1})] - \alpha T \E[\|\nabla_{w_{t-1}} u(w_{t-1})\|_2^2] + \frac{L_1}{2} (\alpha^2 T^2 \E[\| \nabla_{w_{t-1}} u(w_{t-1})\|_2^2] + \sigma_{\frac{t-1}{T}}(f)n) \\
        \leq & ~ \E[u(w_{t-1})] + (\frac{L_1}{2}  \alpha^2 T^2- \alpha T) \E[\|\nabla_{w_{t-1}} u(w_{t-1})\|_2^2] + \frac{L_1}{2}  \sigma_{\frac{t-1}{T}}(f)n 
    \end{align*}
    where the second step follows from Lemma~\ref{lem:tool}, the third step follows from some simple algebras.
\end{proof}

\begin{theorem}[Formal version of Theorem~\ref{thm:main_3}]\label{thm:formal:main_3}
    If the following conditions hold:
    \begin{itemize}
        \item Let $w$ be defined in Definition~\ref{def:w}.
        \item Let $u(w_t)$ be defined in Definition~\ref{def:metric}.
        \item Let $\delta \in (0,0.1)$.
        \item  Let all pre-conditions hold in Lemma~\ref{lem:minimum_func}.
        \item Let $T = \wt{O}(\sqrt{N/(L_1\alpha \epsilon)})$.
        \item Let $1 - \delta$ be the failure probability.
    \end{itemize}
     Then for error $\epsilon > 0$, with a probability at least $1 - \delta$, we have:
    \begin{align*}
        \min_{t \in [T]}\E[\|\nabla_{w_t} u(w_t)\|_2^2] \leq \epsilon
    \end{align*}
\end{theorem}

\begin{proof}
    We have:
    \begin{align}\label{eq:convergence_1}
        \min_{t \in [T]}\E[\|\nabla_{w_t} u(w_t)\|_2^2] \leq & ~ \frac{1}{T} \sum_{t=1}^{T} \E[\|\nabla_{w_t} u(w_t)\|_2^2] \notag \\
        \leq & ~ \frac{1}{\alpha T^2(0.5 L_1 \alpha T-1)} \sum_{t=1}^{T} \E[u(w_{t-1})] - \E[u(w_t)] + \frac{L_1n}{2}\sigma_{\frac{t}{T}}(f) \notag \\ 
        \leq & ~ \frac{1}{\alpha T^2(0.5 L_1 \alpha T-1)} \sum_{t=1}^{T} \E[u(w_{t-1})] - \E[u(w_t)] + \frac{L_1n}{2} \notag \\ 
        \leq & ~ \frac{1}{\alpha T^2(0.5 L_1 \alpha T-1)}(  \E[u(w_{0})] - \E[u(w_T)] + \frac{L_1nT}{2}) 
    \end{align}
    where the first step follows from the minimum is always smaller than the average, the second step follows from Lemma~\ref{lem:decrease}, the third step follows from $\sigma_t(f) \leq 1$, the fourth step follows from simple algebras.

    For the term $\E[u(w_{0})] - \E[u(w_T)]$ in Eq.~\eqref{eq:convergence_1}, we can show that
    \begin{align}\label{eq:convergence_2}
        \E[u(w_{0})] - \E[u(w_T)]
        \leq & ~ \E[u(w_{0})]\notag \\ 
        \leq & ~ O( N\log(N/\delta))
    \end{align}
     where the first step follows from $u(w) \ge 0$ for any $w \in \R^n$, the second step follows from Gaussian tail bound and the upper bound on $f_y$.
    
    Combine Eq.~\eqref{eq:convergence_1} and Eq.~\eqref{eq:convergence_2}, we can show that
    \begin{align*}
        \min_{t \in [T]}\E[\|\nabla_{w_t} u(w_t)\|_2^2] \leq \epsilon
    \end{align*}
    which follows from $T = \wt{O}(\sqrt{N/(L_1\alpha \epsilon)})$.
\end{proof}




\end{document}